\documentclass{article}
\usepackage{ijcai25}

\usepackage{times}
\usepackage{soul}
\usepackage{url}
\usepackage[hidelinks]{hyperref}
\usepackage[utf8]{inputenc}
\usepackage[small]{caption}
\usepackage{amsmath}
\usepackage{amsthm}
\usepackage{algorithm}
\usepackage{algorithmic}
\usepackage[switch]{lineno}
\usepackage{listings}
\usepackage{adjustbox} 
\usepackage{cleveref} 
\usepackage{multirow} 
\usepackage{array}  
\usepackage{color} 
\usepackage{xcolor}
\usepackage{helvet}
\usepackage[T1]{fontenc} 
\usepackage{bbding} 
\usepackage{booktabs} 
\usepackage{tcolorbox}
\tcbuselibrary{breakable}
\usepackage{caption}
\usepackage{courier}
\usepackage{lipsum}
\usepackage[normalem]{ulem} 

\title{How to Mitigate Information Loss in Knowledge Graphs for GraphRAG: Leveraging Triple Context Restoration and Query-Driven Feedback}

\author{
Manzong Huang$^1$ 
\and
Chenyang Bu$^{1*}$\and
Yi He $^2$\and
Xindong Wu$^{1}$\thanks{Corresponding author.}
\affiliations
$^1$ Key Laboratory of Knowledge Engineering with Big Data (the Ministry of Education of China), School of Computer Science and Information Engineering, Hefei University of Technology, China\\
$^2$ School of Computing, Data Sciences, and Physics, College of William and Mary, USA\\
\emails
manzonghuang@mail.hfut.edu.cn,
yihe@wm.edu,
\{chenyangbu, xwu\}@hfut.edu.cn
}

\begin{document}

\maketitle


\begin{abstract}

Knowledge Graph (KG)-augmented Large Language Models (LLMs)
have recently propelled significant advances in complex reasoning tasks, 
thanks to their broad domain knowledge and contextual awareness. 
Unfortunately, current methods often assume KGs to be complete, 
which is impractical 
given the inherent limitations of KG construction and the potential loss of contextual cues when converting unstructured text into entity-relation triples. 
In response,
this paper proposes the Triple Context Restoration and Query-driven Feedback (TCR-QF) framework, which reconstructs the textual context underlying each triple to mitigate information loss, while dynamically refining the KG structure 
by iteratively incorporating query-relevant missing knowledge. 
Experiments on five benchmark question-answering datasets substantiate the effectiveness of TCR-QF in KG and LLM integration, where it
achieves a 29.1\% improvement in Exact Match and a 15.5\% improvement in F1 
over its state-of-the-art GraphRAG competitors.
The code is publicly available at https://github.com/HFUT-DMiC-Lab/TCR-QF.git.

\end{abstract}

\section{Introduction}

Large Language Models (LLMs) augmented with Knowledge Graphs (KGs) have achieved remarkable successes across diverse domains, from social sciences to biomedicine~\cite{roadmap,graph_review,fact_checking_with_kg,biomedical_kg_llm}. 
By harmonizing the structured information in KGs and the sophisticated language understanding and processing capabilities of LLMs, such hybrid systems enable more accurate and context-aware reasoning for complex tasks.


\begin{figure}[!t]
    \centering
    \small
    \includegraphics[width=1.0 \linewidth]{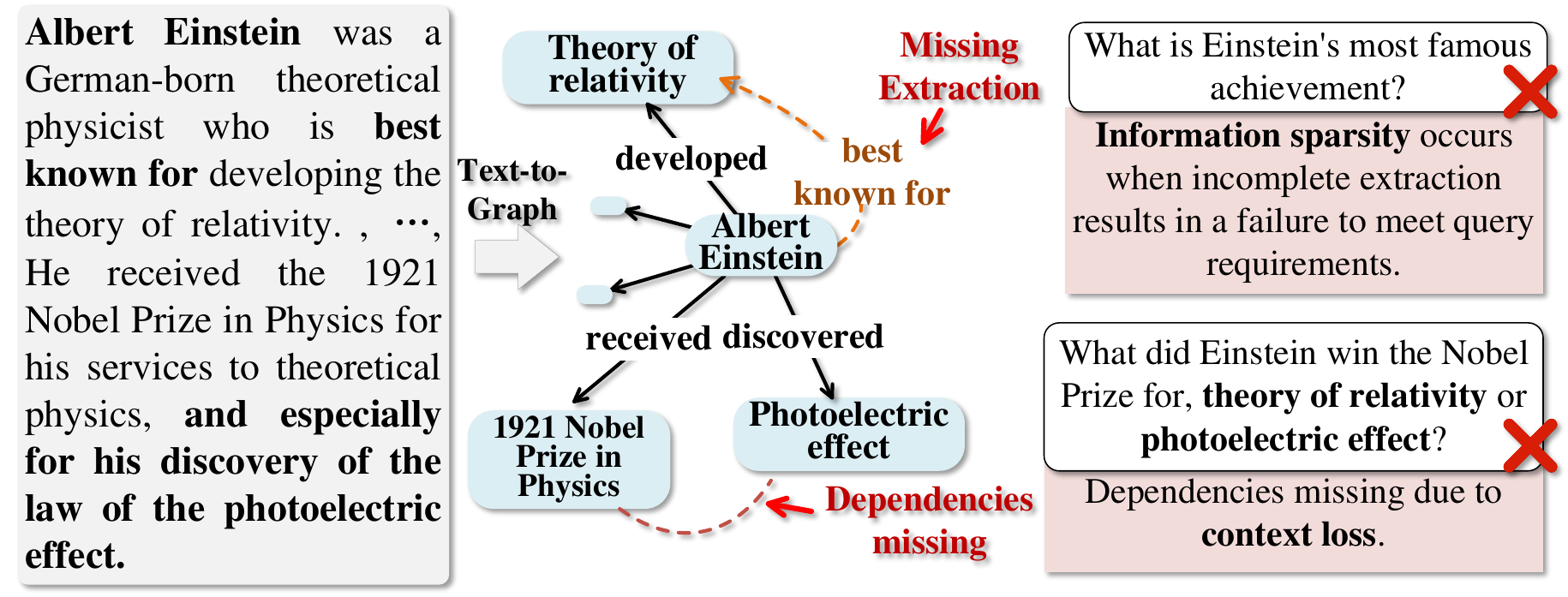}
    \caption{
    Illustration of two main factors of information loss in KGs: \textbf{information sparsity} and \textbf{context loss}. These issues hinder LLMs from accurately answering questions based on KGs.
    }  
    \label{motivation}
    \vspace{-1.5em}
\end{figure}

Despite these advances, the performance of current KG--LLM integration methods is often hindered by the underlying assumption that the KG is complete. 
Typical integration strategy involves retrieving relational data from a constructed KG 
and feed it into LLMs via prompt augmentation~\cite{graph_review,TOG,GraphRAG,G_rag,EWEK},
assuming that critical entities and relationships relevant to the query are already captured within the KG. 
In practice, however, KG construction itself is beset by inherent constraints,
where vital contextual information can be discarded 
in the process of converting unstructured text into structured triples,
leading to missing or incomplete relations~\cite{llm_for_kg,kgc}.
Such missing information 
in KGs can significantly degrade the LLM reasoning capabilities.


To wit, Figure~\ref{motivation} illustrates two primary sources of information loss. 
First, \textbf{information sparsity} arises when information extraction falls short, omitting potentially important triples and thus failing to provide sufficient coverage for specific queries~\cite{incomplete_of_kg,IE_llm,IE_GPT,kg_extract_llm,chen2024sac,sun2024docs2kg,QA-kb}. 
This sparsity can be exacerbated by data noise, long-tail entities, and complex relationships, where extraction algorithms often falter. 
Second, \textbf{context loss} occurs when transforming rich yet unstructured text into discrete triples, sacrificing crucial semantic nuances and relational dependencies~\cite{context_lost_edvi,kg_survey,context_loss_kg}. While prior studies attempt to mitigate this issue by refining graph structures or retrieval algorithms~\cite{KAG,KG_Retriever,HOLMES,munikoti2023atlantic,QA-kb}, their subgraphs still lack the broader contextual information that is vital for robust reasoning, resulting in suboptimal performance in downstream tasks.

To address these challenges, we propose the \textbf{T}riple \textbf{C}ontext \textbf{R}estoration and \textbf{Q}uestion-driven \textbf{F}eedback \textbf{(TCR-QF)} framework,
which aims to restore the missing 
contextual information and dynamically enrich the KG during the reasoning process. 
Specifically, our TCR-QF approach presents a \textit{triple context restoration} component that retrieves the original text passages associated with each triple, thereby recapturing the semantic details often lost during KG construction. We further enhance KG coverage through a \textit{query-driven feedback} mechanism, which iteratively identifies missing information relevant to the query and enriches the KG accordingly. 
These two components together form a synergistic cycle in which contextual fidelity and KG completeness are continuously reinforced, resulting in more accurate and context-aware responses from the LLM.
Empirical study on five benchmark question-answering datasets 
substantiates that TCR-QF significantly outperforms the state-of-the-art GraphRAG methods 
in both response accuracy and completeness,
demonstrating its effectiveness.


\smallskip\noindent
{\bf Specific Contributions} of this paper  are as follows:
\begin{itemize}
\item[1)] We provide a systematic analysis of the key challenges in KG--LLM integration, highlighting the loss of contextual information and incomplete information extraction during KG construction, both of which hinder an advanced LLM reasoning performance. 

\item[2)] We propose the TCR-QF framework, which restores the semantic context associated with triples and employs a query-driven feedback mechanism to iteratively enrich the KG, thereby significantly enhancing the LLM reasoning capabilities. 

\item[3)] Extensive experiments on five benchmark question-answering datasets are carried out, showing that TCR-QF achieves an average 29.1\% improvement in Exact Match and a 15.5\% improvement in F1 over its GraphRAG competitors.
These results validate the merit of restoring contextual information and dynamically updating KGs for effective KG--LLM integration.

\end{itemize}

\section{Related Work}

GraphRAG has emerged as a powerful paradigm for integrating knowledge graphs (KGs) with large language models (LLMs) to advance complex reasoning tasks~\cite{roadmap,graph_review,fact_checking_with_kg,KnowGPT,RATT}. 
A widely adopted strategy involves retrieving relevant subgraphs from preconstructed KGs to augment LLMs during inference~\cite{yasunaga2021qa,taunk2023grapeqa}, with techniques such as extracting hop-$k$ paths around topic entities~\cite{yasunaga2021qa} or focusing on the shortest paths relevant to query entities~\cite{delile2024graph}. More sophisticated methods optimize subgraph retrieval by assigning edge costs~\cite{G_Retriever} or leverage LLMs themselves to generate new relations or invoke function calls~\cite{kim2023kg,jiang2023structgpt}. 

While these approaches have demonstrated effectiveness, most remain limited by their dependence on the initial completeness of the KG and often overlook the contextual information lost during KG construction. 
In reality, KGs are frequently incomplete due to information loss during construction and the difficulties in extracting all relevant triples, especially in noisy or complex scenarios \cite{incomplete_of_kg,QA-kb}. 
These constraints can hinder the ability of LLMs to formulate coherent and context-rich reasoning paths. Addressing these gaps calls for a more dynamic strategy that restores missing contextual details and continuously refines the KG, ensuring that the retrieved and generated knowledge is both accurate and semantically complete.



However, the lack of essential data negatively impacts the inference results of LLMs. To address this, efforts have been made to enhance KG comprehensiveness through refined indexing methods and innovative graph structures for retrieving both triples and texts \cite{KG_Retriever,munikoti2023atlantic,KAG,QA-kb}, as well as using LLMs to improve automated KG construction \cite{kg_extract_llm,IE_llm,IE_GPT}. These methods may retrieve texts related to the query without fully meeting its requirements. Additionally, the retrieved subgraphs can result in the loss of crucial information due to the absence of contextual data within triples, which is essential for maintaining semantic integrity. As a result, the constructed KG may lack critical information necessary for accurate reasoning, leading to suboptimal performance in downstream tasks.

The proposed \textbf{TCR-QF} framework addresses these limitations by dynamically enriching the KG during the reasoning process. By restoring the original textual context of triples, TCR-QF recovers lost semantic information. Additionally, it employs a query-driven feedback mechanism to identify and fill in missing information relevant to a query, enabling the KG to continuously update. This mutual enhancement between KG and LLM improves reasoning performance and better adapts to task requirements.

\section{Proposed Method}

\begin{figure*}
    \centering
    \includegraphics[width=1\linewidth]{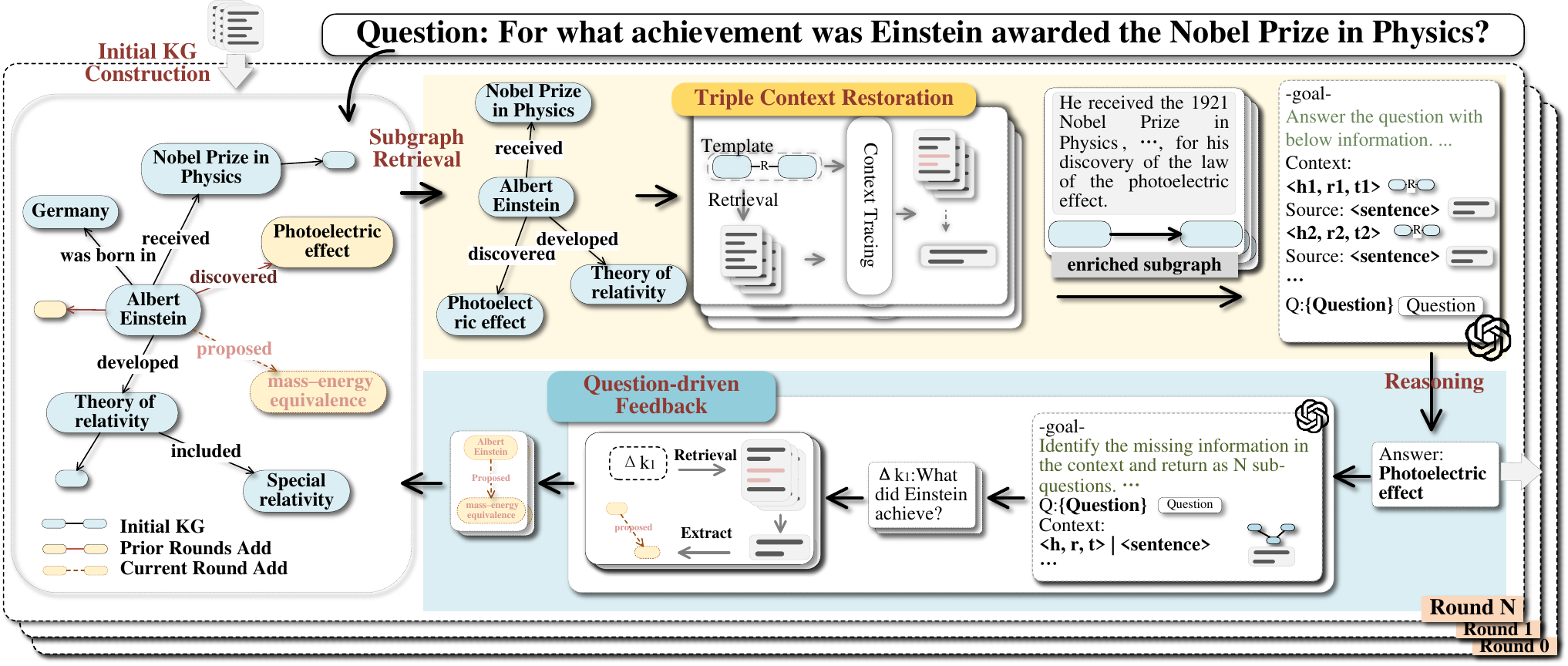}
    \caption{Workflow of TCR-QF. Including a continuously mutual enhancement knowledge flow: (1) Forward Flow: The KG enhances the LLM during answer generation with triple context restoration. (2) Feedback Flow: Identified missing knowledge through query-driven feedback and reinforced into the KG.}
    \label{method}
\vspace{-1em}
\end{figure*}

\noindent
\textbf{Task Definition.}
Given a set of documents \( D = \{d_1, d_2, ..., d_n\} \) and a question \( Q \), the task requires the model to read and reason over multiple relevant documents, extract and aggregate the necessary information, and finally generate the answer \( A \) based on the information from the documents.

In this section, we present the \textbf{TCR-QF} framework, designed to mitigate the loss of contextual information when building knowledge graphs (KGs) from unstructured text and to dynamically enrich these graphs during the reasoning process. 
As shown in Figure~\ref{method}, the framework comprises four key components: 
\textbf{(1) Knowledge Graph Construction}, which builds a unified KG from textual sources; 
\textbf{(2) Subgraph Retrieval}, responsible for extracting task-relevant subgraphs composed of potential reasoning paths; 
\textbf{(3) Triple Context Restoration}, which traces back the original textual context of each triple to recover lost semantic nuances; and 
\textbf{(4) Iterative Reasoning with Query-Driven Feedback}, where an iterative cycle that both generates answers and identifies missing knowledge, thereby refining the KG on-the-fly. Together, these components ensure that contextual details are preserved and the KG remains up-to-date, ultimately enhancing the quality and depth of the system reasoning.

The above steps establishes a synergistic cycle for two-way knowledge enhancement, namely,
\paragraph{Forward Flow:} The KG informs the LLM during answer generation, represented as
  \[
  G^{(i)} \longrightarrow G'^{(i)} \longrightarrow A_{\text{LLM}}^{(i)}. 
  \]
In the \(i\)-th iteration, a subgraph is retrieved from the KG \( G^{(i)} \) and enhanced through triple context restoration to form \( G'^{(i)} \). The LLM then infers the answer \( A_{\text{LLM}}^{(i)} \) based on \( G'^{(i)} \).
\paragraph{Feedback Flow:} The missing knowledge in the KG is identified and subsequently integrated back into the KG:
  \[
   G^{(i+1)} \longleftarrow \Delta K^{(i)} \longleftarrow A_{\text{LLM}}^{(i)},
  \]
where \( \Delta K^{(i)} \) represents the knowledge increment  corresponding to the missing knowledge. This increment is updated in the KG as triples, resulting in the more comprehensive KG  \( G^{(i+1)} \).


\subsection{Knowledge Graph Construction}
An initial KG was constructed from raw textual data using LLMs to extract entities and relations as triples \( (e_h, r, e_t) \), where entities include names, types \( \text{Type}(e) \), and descriptions \( \text{Desc}(e) \). Source document information is retained in each node for provenance. The construction involves:

\paragraph{Document Splitting}
Each document \( D \) of length \( L \) is divided into overlapping chunks \( C_i \) with maximum length \( \text{MAX\_LEN} = 512 \) and overlap \( \text{OVERLAP} = 64 \) tokens:
\[
\begin{aligned}
C_i &= D[s_i : e_i], \\
s_i &= (i - 1) \times (\text{MAX\_LEN} - \text{OVERLAP}) + 1, \\
e_i &= \min(s_i + \text{MAX\_LEN} - 1, L).
\end{aligned}
\]
The overlap ensures entities and relations spanning across chunks are captured.

\paragraph{Triple Extraction}
From each chunk \( C_i \), the LLM extracts triples:
\[
T_i = \text{ExtractTriples}(C_i),
\]
where \( T_i \) is the set of triples from \( C_i \). A specialized prompt guides the LLM to output structured information, including entity types and descriptions.


\subsection{Subgraph Retrieval}
The subgraph retrieval phase focuses on extracting pertinent information from the KG in response to the query. Specifically, given a query \( Q \) expressed in natural language, the retrieval stage aims to extract the most relevant elements (e.g., entities, triplets, paths, subgraphs) from KGs, which can be formulated as:
\[
G^* = \text{G-Retriever}(Q, G)
\]
\[
= \arg \max_{G \subseteq \mathcal{R}(G)} \text{Sim}(Q, G),
\]
where \( G^* = \{ (h_0, r_0, s_1), (h_1, r_1, s_2), \dots, (h_t, r_t, s_{t+1}) \}\) is the optimal retrieved graph elements and \( \text{Sim}(\cdot, \cdot) \) is a function that measures the semantic similarity between user queries and the graph data. \( \mathcal{R}(G) \) represents a function to narrow down the search range of subgraphs, considering efficiency. The retrieval method employed in the TCR-QF builds upon existing KG retrieval method~\cite{TOG}, which utilize LLMs to perform a beam search over the KG, with iterative pruning guided by the LLM.

\subsection{Triple Context Restoration}
The structuring of unstructured text into triples can lead to a loss of semantic context. To address this issue, a triple context restoration mechanism was implemented in TCR-QF to restore semantic integrity by tracing back to the original textual context of the triples.
\paragraph{Context Retrieval}
For each triple \( (e_h, r, e_t) \) in the retrieved subgraphs, the source documents associated with the head and tail entities were retrieved:
\[
\text{Sources}_{(e_h, e_t)} = \text{Sources}(e_h) \cup \text{Sources}(e_t).
\]
These sources were the documents which the entities were originally extracted during KG construction. This set encompassed all documents potentially containing contextual information about the relationship between \( e_h \) and \( e_t \).

\paragraph{Triple Context Tracing}
To trace the context of the triple \( (e_h, r, e_t) \), the most relevant sentence from source documents were identified. A template \( T_{(e_h, r, e_t)} \) was used, such as:
\[
T_{(e_h, r, e_t)} = \text{``} e_h \text{ } r \text{ } e_t \text{''}.
\]
A pretrained embedding model \( f_{\text{embed}} \) was used to generate embeddings for both the template and candidate sentences. The context relevance was assessed via cosine similarity:
\[
\mathbf{v}_T = f_{\text{embed}}(T_{(e_h, r, e_t)}), \mathbf{v}_s = f_{\text{embed}}(s), \quad \forall s \in \mathcal{S},
\]
\[
\text{sim}(\mathbf{v}_T, \mathbf{v}_s) = \frac{\mathbf{v}_T^\top \mathbf{v}_s}{\|\mathbf{v}_T\| \cdot \|\mathbf{v}_s\|}
\]

where \( \mathcal{S} \) is the set of all sentences extracted from \( \text{Sources}_{(e_h, e_t)} \). The sentence with the highest similarity score was selected to provide contextual information into the triple. 
\paragraph{Triple Augmentation}
Each triple was augmented with its associated contextual sentence:
\[
(e_h, r, e_t) \longrightarrow (e_h, r, e_t, \mathcal{S}_{\text{top}}).
\]
This augmentation restored the contextual information of the triples, improving the accuracy and depth of inference tasks that rely on the KG.

\subsection{Iterative Reasoning with Query-driven Feedback}

To generate accurate answers to the original queries \( Q \), an iterative reasoning process incorporating a query-driven feedback mechanism was implemented. This approach dynamically enriches the KG by identifying and updating missing information during the reasoning process, thereby enhancing the LLM‘s capability to produce more accurate responses.

Initially, the enriched subgraph \( G'^{(0)} \) obtained from triple context restoration was used to prompt the LLM:
\[
\mathcal{I}^{(0)} = \text{FormatInput}(Q, G'^{(0)}).
\]
The LLM then generated an initial answer \( A_{\text{LLM}}^{(0)} \) by processing this prompt:

\[
A_{\text{LLM}}^{(0)} = \text{LLM\_Generate}(\mathcal{I}^{(0)}),
\]

where \( \text{LLM\_Generate} \) refers to generating a response based on the formatted input \( \mathcal{I}^{(0)} \).

\paragraph{Missing Knowledge Identification}

The initial answer and contexts were analyzed to identify missing information required for the query:

\[
\Delta K^{(0)} = \text{IdentifyMissing}(Q, A_{\text{LLM}}^{(0)}, G'^{(0)}),
\]

where \( \Delta K^{(0)} \) represents the set of missing knowledge, formalized as a series of sub-questions. The function \( \text{IdentifyMissing} \) utilizes the LLM to compare \( Q \) with \( A_{\text{LLM}}^{(0)} \) and \( G'^{(0)} \), effectively harnessing its understanding to identify gaps in knowledge.

\paragraph{Knowledge Graph Enrichment}

For each missing component \( k_q \in \Delta K^{(0)} \), a dense retriever interacted with the original text sources \( \mathcal{D} \) to retrieve relevant textual information and extract the missing knowledge:
\[
\mathcal{D}_{\text{relevant}} = \text{DenseRetrieve}(k_q, \mathcal{D}),
\]
\[
k = \text{ExtractTriples}(k_q, \mathcal{D}_{\text{relevant}}),
\]
where \( \text{ExtractTriples} \) employs the LLM to find and extract the needed information, resulting in triples \( k \) corresponding to the missing knowledge. The KG was then updated:
\[
G^{(1)} = G^{(0)} \cup \Delta K^{(0)} \quad \text{with} \quad \forall~k \in \Delta K^{(0)}, \, k \notin G^{(0)}.
\]
Duplicate relationships were filtered based on edit distance from elements in \( G^{(0)} \) to maintain uniqueness in \( G^{(1)} \). A dense passage retriever, implemented using OpenAI's \texttt{text-embedding-small}, was employed due to its effectiveness in retrieving semantically relevant passages.

\paragraph{Iterative Reasoning and Update}
The updated KG \( G^{(1)} \) was used to generate a new answer by following the reasoning steps:
\[
A_{\text{LLM}}^{(1)} = \text{LLM\_Generate}(\text{FormatInput}(Q, G^{(1)})).
\]
This iterative process continued, repeating the steps of Missing Knowledge Identification and Knowledge Graph Enrichment:
\[
\begin{aligned}
\Delta K^{(i)} &= \text{IdentifyMissing}(Q, A_{\text{LLM}}^{(i-1)}, G^{(i-1)}), \\
G^{(i)} &= G^{(i-1)} \cup \Delta K^{(i)}, \\
A_{\text{LLM}}^{(i)} &= \text{LLM\_Generate}(\text{FormatInput}(Q, G^{(i)})),
\end{aligned}
\]

for \( i = 2, 3, \dots \), until \( \Delta K^{(i)} = \emptyset \) or a predefined maximum number of iterations \( I_{\text{max}} = 20 \) was reached. By analyzing retrieved contexts and generated responses at each iteration, gaps in the KG were detected and addressed, continuously optimizing the KG and enhancing the reasoning capabilities of the LLM.

Due to space constraints, the detailed prompts used for the LLM at each step are provided in the appendix\footnote{The appendix is available in the arXiv version: \url{https://arxiv.org/abs/2501.15378}.}.

\section{Experiments}
To evaluate the effectiveness of the TCR-QF on question-answering tasks, experiments were conducted on 5 question-answering datasets: 2WikiMultiHopQA~\cite{2WikiMultiHopQA}, HotpotQA~\cite{HotpotQA}, ConcurrentQA~\cite{ConcurrentQA}, MuSiQue-Ans and MuSiQue-Full~\cite{musique} . Followed the settings outlined in ~\cite{HotpotQA}, utilizing a collection of related contexts for each pair as the retrieval corpus. Exact Match (EM) and F1 score were presented as the evaluation metrics across all datasets. 

We compared TCR-QF with representative methods from LLMs and RAG:\\
\textbf{(1) LLM Only}: Methods that directly use LLMs for obtaining answers, including models such as \texttt{gpt-4o-mini} and \texttt{gpt-4o}, as well as chain-of-thought (CoT)~\cite{wei2022chain} prompting strategies.\\
\textbf{(2) Text-based RAG}: Methods that employ a dense retriever to retrieve relevant text chunks from a text corpus and generate answers by leveraging this information. For this category, LangChainQ\&A\footnote{\url{https://python.langchain.com/docs/tutorials/rag}} was used as a representative naive RAG method, which is well-known and widely used.\\
\textbf{(3) Graph-based RAG}: Methods that retrieve subgraphs from KG to enhance LLM. ToG~\cite{TOG} was selected as a representative for comparison in this category.\\
\textbf{(4) Hybrid RAG}: Methods like GraphRAG~\cite{GraphRAG} that retrieve information from both KG and textual documents to augment LLM.

\paragraph{Experimental Settings:}

For all comparison methods and the TCR-QF, unless otherwise specified, the \texttt{gpt-4o-mini-2024-07-18} model was utilized. Due to the high computational costs associated with inference on the full dataset, 1,200 samples were randomly selected from each of the larger datasets—2WikiMultiHopQA, HotpotQA, MuSiQue-Full, and MuSiQue-Ans—for testing to conserve computational resources. For ConcurrentQA, 1,600 samples from the complete test set were evaluated.

\subsection{Results and Findings}

Table~\ref{tab:main_result} presents the comparative results,
from which we answer the following Research Question (RQ).

\begin{table*}[ht]
\centering
\small
\begin{tabular}{cccccccccccc}
\toprule
\multirow{2}{*}{\textbf{Method Type}} & \multirow{2}{*}{\textbf{Method}} & \multicolumn{2}{c}{\textbf{2WikiMultiHopQA}} & \multicolumn{2}{c}{\textbf{HotpotQA}} & \multicolumn{2}{c}{\textbf{MuSiQue-Full}} & \multicolumn{2}{c}{\textbf{MuSiQue-Ans}} & \multicolumn{2}{c}{\textbf{ConcurrentQA}} \\
\cmidrule(lr){3-4} \cmidrule(lr){5-6} \cmidrule(lr){7-8} \cmidrule(lr){9-10} \cmidrule(lr){11-12}
& & \textbf{EM} & \textbf{F1} & \textbf{EM} & \textbf{F1} & \textbf{EM} & \textbf{F1} & \textbf{EM} & \textbf{F1} & \textbf{EM} & \textbf{F1} \\
\midrule
\multirow{4}{*}{\textbf{LLM only}} & \textbf{GPT-4o-mini} & 0.266 & 0.320 & 0.273 & 0.381 & 0.048 & 0.132 & 0.052 & 0.135 & 0.112 & 0.178 \\
\cmidrule(lr){2-12}
& \textbf{GPT-4o} & 0.311 & 0.364 & 0.351 & 0.475 & 0.089 & 0.193 & 0.104 & 0.215 & 0.176 & 0.247 \\
\cmidrule(lr){2-12}
& \textbf{CoT} & 0.287 & 0.354 & 0.299 & 0.420 & 0.093 & 0.196
 & 0.117 & 0.219 & 0.134 & 0.203 \\
\midrule
\textbf{Text-based} & \textbf{Naive RAG} & 0.339 & 0.391 & 0.411 & 0.530 & 0.111 & 0.207 & 0.122 & 0.221 & 0.363 & 0.443 \\
\midrule
\textbf{Graph-based} & \textbf{TOG} & 0.400 & 0.476 & 0.420 & 0.555 & 0.136 & 0.237 & 0.160 & 0.269 & 0.278 & 0.359 \\
\midrule
\textbf{Hybrid} & \textbf{GraphRAG} & 0.485 & 0.626 & 0.495 & 0.645 & 0.189 & 0.326 & 0.258 & 0.395 & 0.459 & 0.582 \\
\midrule
\textbf{Proposed} & \textbf{TCR-QF} & \textbf{0.598} & \textbf{0.680} & \textbf{0.558} & \textbf{0.708} & \textbf{0.303} & \textbf{0.432} & \textbf{0.366} & \textbf{0.489} & \textbf{0.492} & \textbf{0.597} \\
\bottomrule
\end{tabular}

\caption{ Main Results. Performance comparison of different methods across five question answering datasets.}
\vspace{-1em}
\label{tab:main_result}
\end{table*}

\begin{description}
    \item[RQ1]  {\em How does the TCR-QF improve the completeness and accuracy of information retrieval in question answering tasks compared to the existing GraphRAG methods?}
\end{description}

Table~\ref{tab:main_result} demonstrates the superiority of the TCR-QF compared to different methods on five benchmark question answering datasets. TCR-QF consistently achieves the highest EM and F1 scores across all datasets, demonstrating its superior effectiveness in enhancing LLMs for complex reasoning tasks.
Compared to the \textbf{LLM-only} approaches (\textbf{GPT-4o-mini}, \textbf{GPT-4o} and \textbf{CoT}), TCR-QF shows substantial improvements. For instance, on the HotpotQA dataset, TCR-QF attains an EM score of 0.558, which is \textbf{0.207} higher than GPT-4o's score of 0.351, representing a relative improvement of approximately \textbf{59\%}. This indicates that while LLMs possess strong language understanding capabilities, integrating external knowledge as TCR-QF does markedly enhances their accuracy in answering complex questions.

When contrasting TCR-QF with the \textbf{text-based} method (\textbf{Naive RAG}) and the \textbf{graph-based} method (\textbf{ToG}), TCR-QF exhibits notable performance gains. Specifically, on the 2WikiMultiHopQA dataset, TCR-QF achieves an EM score of \textbf{0.598}, which is an absolute increase of \textbf{0.259} over Naive RAG's EM score of \textbf{0.339}—a relative improvement of approximately \textbf{76.4\%}. Similarly, TCR-QF surpasses TOG's EM score of \textbf{0.400} by an absolute margin of \textbf{0.198}, reflecting a \textbf{49.5\%} improvement. This significant enhancement indicates that TCR-QF's approach of enriching the LLM with more comprehensive knowledge markedly improves reasoning, outperforming methods that rely solely on retrieved texts or static KGs.

Furthermore, TCR-QF outperforms the \textbf{hybrid} method (\textbf{GraphRAG}), which combines text and graph information. On the MuSiQue-Full dataset, TCR-QF achieves an EM score of \textbf{0.303}, compared to GraphRAG's EM score of \textbf{0.189}. This represents an absolute increase of \textbf{0.114}, amounting to an improvement of approximately \textbf{60.3\%}. These significant gains demonstrate that TCR-QF effectively leverages knowledge to enhance the LLM's performance beyond what is achieved by simply combining text and graph data. By dynamically restoring lost semantic information and enriching the KG during reasoning, TCR-QF provides a more comprehensive context for the LLM, leading to better reasoning and answer generation in complex tasks.

The consistent superiority of TCR-QF across multiple datasets—ranging from general question-answering to those requiring multi-hop reasoning—highlights TCR-QF's robustness and general applicability. TCR-QF effectively addresses the challenges posed by incomplete KGs and information loss, leading to more accurate and complete responses. 

\subsection{Ablation Study}

\begin{table}[htbp]
\centering
\small
\begin{tabular}{llcccc}
\toprule
\multirow{2}{*}{\textbf{Methods}} & \multicolumn{2}{c}{\textbf{2WikiMultiHopQA}} & \multicolumn{2}{c}{\textbf{HotpotQA}} \\
\cmidrule(lr){2-3} \cmidrule(lr){4-5}
& \textbf{EM} & \textbf{F1} & \textbf{EM} & \textbf{F1} \\
\midrule
\textbf{ToG(w/o TCR\&QF)} & 0.400 & 0.476 & 0.420 & 0.555 \\
\textbf{TCR(w/o QF)} & 0.481 & 0.561 & 0.494 & 0.642 \\
\textbf{QF(w/o TCR)} & 0.568 & 0.651 & 0.515 & 0.656 \\
\textbf{TCR-AF} & 0.538 & 0.619 & 0.531 & 0.682 \\
\textbf{TCR-QF} & \textbf{0.598} & \textbf{0.680} & \textbf{0.558} & \textbf{0.708} \\
\bottomrule
\end{tabular}
\caption{Ablation experiment results on the 2WikiMultiHopQA and HotpotQA datasets. \textbf{TCR} stands for triple context restoration, \textbf{QF} stands for query-driven feedback. \textbf{TCR-AF} indicates replacing query-driven feedback with answer-driven feedback which directly extract triples from the answers and feed them back into the KG.}
\label{tab:ablation}
\vspace{-1em}
\end{table}

To evaluate the individual contributions of the proposed components, namely 
\textit{triple context restoration} (TCR) and \textit{query-driven feedback} (QF), to the overall performance of the TCR-QF, an ablation study was conducted on the 2WikiMultiHopQA and HotpotQA datasets to answer the question: 

\begin{description}
    \item[RQ2]  {\em In what ways does each component in TCR-QF enhance the reasoning of the LLM?}
\end{description}


Table~\ref{tab:ablation} presents the results of the ablation experiments. The full \textbf{TCR-QF} is compared with several ablated variants:
\begin{itemize}
    \item \textbf{ToG (w/o TCR \& QF)}: The baseline method operating on the KG.
    \item \textbf{TCR (w/o QF)}: Incorporates triple context restoration alone to address contextual information loss.
    \item \textbf{QF (w/o TCR)}: Employs query-driven feedback alone to approach incomplete information extraction.
    \item \textbf{TCR-AF}: Integrate triple context restoration with \textit{answer-driven feedback} (AF) which involves directly extracting triples from the LLM's answer and adding them to the KG.
\end{itemize}

From the results we can draw the following insights. 

\paragraph{Effectiveness of Triple Context Restoration (TCR).} Comparing the baseline \textbf{ToG} method with the \textbf{TCR} variant, it can be observed that introducing triple context restoration leads to significant performance improvements. On the 2WikiMultiHopQA dataset, the EM score increases from 0.400 to 0.481, representing an improvement of 20.25\%, while the F1 score rises from 0.476 to 0.561. Similarly, on the HotpotQA dataset, the EM score improves from 0.420 to 0.494 (a 17.62\% improvement), and the F1 score increases from 0.555 to 0.642. These enhancements confirm that triple context restoration effectively mitigates contextual information loss by reconnecting structured triples with their original textual context, thereby enriching the semantic information available for reasoning.

\paragraph{Effectiveness of Query-Driven Feedback (QF).} The \textbf{QF} variant, which focuses on dynamically updating the KG based on the requirements of the query, shows even greater improvements over the baseline. The EM scores rise to 0.568 (a 42.00\% improvement) on 2WikiMultiHopQA and 0.522 (a 24.29\% improvement) on HotpotQA. These substantial gains indicates that query-driven feedback significantly addresses the issue of incomplete information extraction. By dynamically enriching the KG based on the specific requirements of the query, the model fills in the missing knowledge that static KGs might overlook due to limitations in initial extraction algorithms. This adaptive approach continually enhances the relevance and comprehensiveness of the knowledge graph throughout the reasoning process.

\paragraph{Synergy of TCR and QF.} The full \textbf{TCR-QF} method, which combines both triple context restoration and query-driven feedback, achieves the highest performance. EM scores reach 0.598 on 2WikiMultiHopQA and 0.558 on HotpotQA, with relative improvements of 49.50\% and 32.86\% over the baseline, respectively. These results underscore a synergistic effect when combining TCR and QF, as the model benefits from both restored contextual semantics and a dynamically enriched KG. The integration of both components effectively addresses the dual challenges of information loss, leading to more accurate and complete reasoning.

\paragraph{Comparison with Answer-Driven Feedback (TCR-AF).} The \textbf{TCR-AF} variant replaces query-driven feedback with answer-driven feedback, where triples are extracted from the model's answers to update the KG. While TCR-AF outperforms \textbf{ToG}, achieving EM scores of 0.538 on 2WikiMultiHopQA and 0.523 on HotpotQA, it falls short compared to \textbf{TCR-QF}. TCR-QF scores 0.598 on 2WikiMultiHopQA, an 11.12\% increase over TCR-AF. This suggests that enriching the KG proactively based on the query is more effective than reactively updating it based on answers, likely because it prevents error propagation from incomplete initial reasoning.

\begin{figure}
    \centering
    \includegraphics[width=1.0\linewidth]{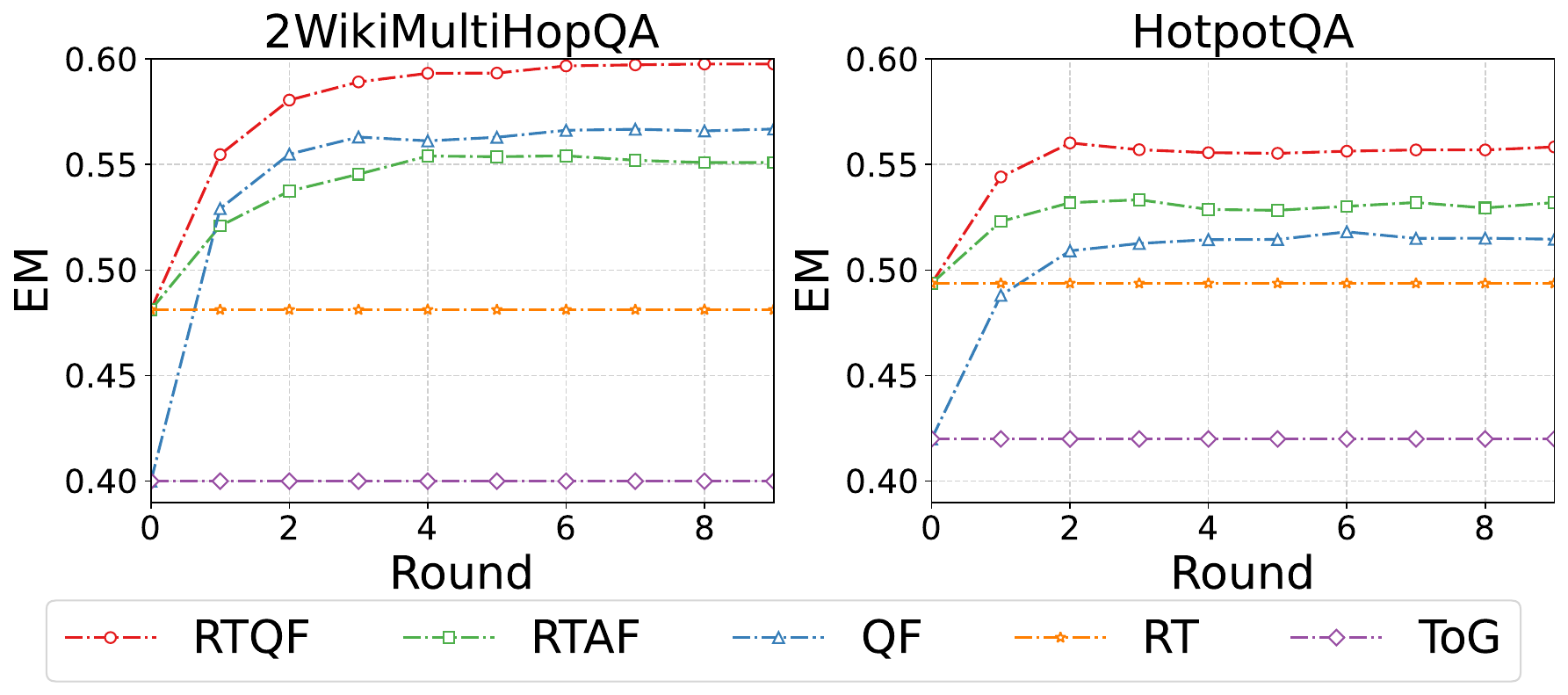}
    \caption{Comparative ressults from the ablation study. EM performance of different methods across rounds on 2WikiMultiHopQA and HotpotQA.}
    \label{fig:metric_trends}
    \vspace{-1em}
\end{figure}
\paragraph{Performance Trends Across Rounds.} 
Figure~\ref{fig:metric_trends} illustrates the EM performance across multiple reasoning rounds for each method. It is evident that \textbf{TCR-QF} consistently outperforms other variants from the initial rounds and maintains its lead as the reasoning progresses. The performance gain from TCR-QF is additive, with TCR-QF achieving the highest accuracy. The diminishing returns after a few rounds indicate that the most significant knowledge enrichment occurs early in the reasoning process, emphasizing the performance of proposed method.

In conclusion, the ablation study corroborates our initial hypotheses, demonstrating that both \textit{triple context restoration} and \textit{query-driven feedback} are vital in addressing the inherent limitations of integrating KGs with LLMs. Individually, each component contributes significantly to performance improvements by targeting specific sources of information loss—triple context restoration restores essential contextual semantics lost during the structuring process, while query-driven feedback dynamically enriches the KG to address incomplete information extraction. These results highlight the effectiveness of restoring semantic integrity and continuously updating the KG during reasoning, fulfilling our research objectives and underscoring the importance of a bidirectional knowledge flow in optimizing reasoning outcomes.

\subsection{Statistical and Convergence Analysis}

To evaluate the effectiveness and convergence of the TCR-QF, statistical analyses were conducted over multiple inference rounds. Table~\ref{tab:statistic_on_nodes_edges} presents key metrics from the initial round to the 10th round, including the numbers of nodes and edges in the KG, as well as the EM and F1 scores on the 2WikiMultiHopQA dataset.
These experiments and results provide answers to the following question: 

\begin{description}
    \item[RQ3]  {\em How do TCR-QF continuously enhance KG and boost LLM reasoning?}
\end{description}


\begin{table}[!ht]
\centering
\small
\begin{tabular}{c c c c c}
\toprule
 & \multicolumn{4}{c}{\textbf{2WikiMultiHopQA}} \\
\cmidrule(lr){2-5}
\textbf{Rounds} & \textbf{Nodes} & \textbf{Edges} & \textbf{EM} & \textbf{F1} \\
\midrule
\textbf{0} & 74,571 & 69,866 & 0.481 & 0.562 \\
\textbf{1} & 76,441 & 74,006 & 0.555 & 0.637 \\
\textbf{2} & 77,377 & 76,615 & 0.581 & 0.662 \\
\textbf{3} & 77,937 & 78,265 & 0.589 & 0.671 \\
\textbf{4} & 78,259 & 79,258 & 0.593 & 0.676 \\
\textbf{5} & 78,450 & 79,840 & 0.593 & 0.675 \\
\textbf{6} & 78,570 & 80,150 & 0.597 & 0.679 \\
\textbf{7} & 78,630 & 80,310 & 0.597 & 0.679 \\
\textbf{8} & 78,650 & 80,403 & 0.598 & 0.680 \\
\textbf{9} & 78,656 & 80,446 & 0.598 & 0.680 \\
\textbf{10} & 78,661 & 80,472 & 0.598 & 0.680 \\
\cmidrule(lr){1-5}
\textbf{$\Delta$} & 4,090 & 10,606 & 0.117 & 0.118 \\
\bottomrule
\end{tabular}
\caption{Statistics from the initial round to the 10th round on 2WikiMultiHopQA dataset, where $\Delta$ denotes the cumulative increase.}

\label{tab:statistic_on_nodes_edges}
\end{table}

From the results we can draw the following insights. 

\textbf{Continuous Improvement of KG Completeness and Model Reasoning Performance.} As demonstrated in Table~\ref{tab:statistic_on_nodes_edges}, the TCR-QF significantly enriches the KG over successive inference rounds. Specifically, on the 2WikiMultiHopQA dataset, the number of nodes in the KG increased by 4,090 (from 74,571 to 78,661), and the number of edges increased by 10,606 (from 69,866 to 80,472) over 10 rounds. This enrichment directly addresses the issue of information sparsity by incorporating previously missing triples and expanding the KG's coverage to meet query demands. Correspondingly, the model's reasoning performance improved substantially. The Exact Match (EM) score increased from 0.481 to 0.598, a 24.3\% improvement, and the F1 score rose from 0.562 to 0.680, a 21.0\% improvement. These significant performance gains indicate that the enriched KG provides the LLM with more comprehensive and contextually rich information, directly mitigating the effects of context loss and enhancing reasoning accuracy.

\begin{figure}
    \centering
    \includegraphics[width=1.0\linewidth]{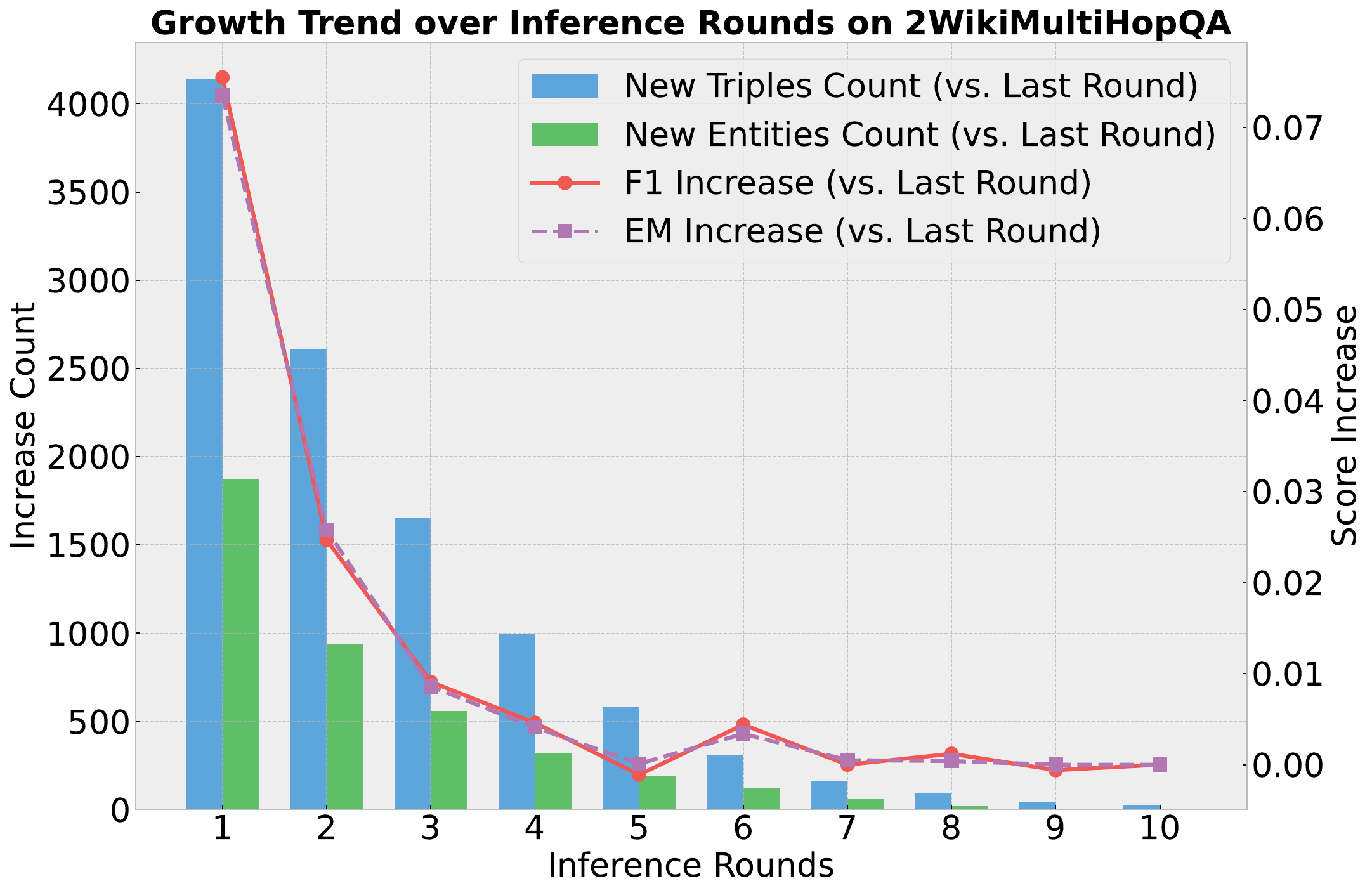}
    \caption{Trends in KG growth and inference performance improvement over rounds on 2WikiMultiHopQA.}
    \label{fig:growth_trend}
    \vspace{-1em}
\end{figure}

\textbf{Alignment of KG Completeness and Reasoning Performance Enhancement.} As depicted in Figure~\ref{fig:growth_trend}, the parallel upward trends in KG metrics and performance scores affirm a strong correlation between the enriched KG and the model's improved reasoning ability. By restoring the contextual information associated with triples and integrating new, relevant knowledge through query-driven feedback, the TCR-QF enhances the semantic integrity of the KG. This comprehensive knowledge base enables the LLM to perform more accurate and context-aware reasoning, directly addressing the limitations posed by information sparsity and context loss.

\textbf{Convergence of KG Enrichment and Performance Improvements.} The TCR-QF not only enriches the KG but also exhibits convergence over inference rounds, ensuring efficient use of computational resources. As illustrated in Figure~\ref{fig:growth_trend}, both the growth of the KG and the improvement in performance metrics begin to plateau after several rounds, specifically between the 8th and 10th iterations. The incremental increases in nodes and edges diminish, and the EM and F1 scores stabilize at 0.598 and 0.680, respectively. This convergence suggests that the TCR-QF effectively enriches the KG to an optimal level, beyond which additional iterations yield minimal benefits.

The experimental results validate the effectiveness of the TCR-QF in overcoming the foundational challenges outlined in the introduction. By continuously and efficiently enhancing the KG's completeness and restoring lost contextual nuances, the TCR-QF significantly boosts the model's reasoning performance. These findings confirm that addressing information loss through dynamic KG enrichment and context restoration is a viable and efficient strategy for advancing the integration of KGs and LLMs in complex reasoning tasks.
\section{Conclusion}
This paper introduces \textbf{TCR-QF}, a novel framework that integrates knowledge graphs (KGs) with large language models (LLMs) to enhance complex question answering. By mitigating \emph{context loss} through triple context restoration (TCR) and addressing \emph{incomplete extraction} with query-driven feedback (QF), TCR-QF recovers key semantic details and dynamically expands the KG during reasoning. Experiments on five benchmarks show that TCR-QF outperforms state-of-the-art methods, demonstrating the benefits of contextualized triples and iterative KG updates. These results highlight the potential of TCR-QF to bridge the gap between structured and unstructured knowledge, paving the way for more accurate and robust AI-driven reasoning across diverse domains.
 
\section*{Acknowledgements}
This work was supported by the National Natural Science Foundation of China (Grant No. 62120106008), the Anhui Provincial Science and Technology Fortification Plan (Grant No. 202423k09020015), and the Youth Talent Support Program of the Anhui Association for Science and Technology (Grant No. RCTJ202420). The authors gratefully acknowledge these supports.

\bibliographystyle{named}
\bibliography{ijcai25}

\begin{thebibliography}{}

\bibitem[\protect\citeauthoryear{Arora \bgroup \em et al.\egroup }{2023}]{ConcurrentQA}
Simran Arora, Patrick S.~H. Lewis, Angela Fan, Jacob Kahn, and Christopher R{\'{e}}.
\newblock Reasoning over public and private data in retrieval-based systems.
\newblock {\em Trans. Assoc. Comput. Linguistics}, 11:902--921, 2023.

\bibitem[\protect\citeauthoryear{Biswas \bgroup \em et al.\egroup }{2024}]{incomplete_of_kg}
Russa Biswas, Harald Sack, and Mehwish Alam.
\newblock {MADLINK:} attentive multihop and entity descriptions for link prediction in knowledge graphs.
\newblock {\em Semantic Web}, 15(1):83--106, 2024.

\bibitem[\protect\citeauthoryear{Chen \bgroup \em et al.\egroup }{2024a}]{chen2024sac}
Hanzhu Chen, Xu~Shen, Qitan Lv, Jie Wang, Xiaoqi Ni, and Jieping Ye.
\newblock {SAC-KG:} exploiting large language models as skilled automatic constructors for domain knowledge graphs.
\newblock {\em CoRR}, abs/2410.02811, 2024.

\bibitem[\protect\citeauthoryear{Chen \bgroup \em et al.\egroup }{2024b}]{KG_Retriever}
Weijie Chen, Ting Bai, Jinbo Su, Jian Luan, Wei Liu, and Chuan Shi.
\newblock Kg-retriever: Efficient knowledge indexing for retrieval-augmented large language models.
\newblock {\em CoRR}, abs/2412.05547, 2024.

\bibitem[\protect\citeauthoryear{Cohen \bgroup \em et al.\egroup }{2023}]{QA-kb}
William~W. Cohen, Wenhu Chen, Michiel de~Jong, Nitish Gupta, Alessandro Presta, Pat Verga, and John Wieting.
\newblock {QA} is the new {KR:} question-answer pairs as knowledge bases.
\newblock In {\em Proceedings of 37th Conference on AAAI}, pages 15385--15392. {AAAI} Press, 2023.

\bibitem[\protect\citeauthoryear{Dehghan \bgroup \em et al.\egroup }{2024}]{EWEK}
Mohammad Dehghan, Mohammad~Ali Alomrani, Sunyam Bagga, David Alfonso{-}Hermelo, Khalil Bibi, Abbas Ghaddar, Yingxue Zhang, Xiaoguang Li, Jianye Hao, Qun Liu, Jimmy Lin, Boxing Chen, Prasanna Parthasarathi, Mahdi Biparva, and Mehdi Rezagholizadeh.
\newblock {EWEK-QA} : Enhanced web and efficient knowledge graph retrieval for citation-based question answering systems.
\newblock In {\em Proceedings of the 62nd {ACL} 2024}, pages 14169--14187. ACL, 2024.

\bibitem[\protect\citeauthoryear{Delile \bgroup \em et al.\egroup }{2024}]{delile2024graph}
Julien Delile, Srayanta Mukherjee, Anton~Van Pamel, and Leonid Zhukov.
\newblock Graph-based retriever captures the long tail of biomedical knowledge.
\newblock {\em CoRR}, abs/2402.12352, 2024.

\bibitem[\protect\citeauthoryear{Edge \bgroup \em et al.\egroup }{2024}]{GraphRAG}
Darren Edge, Ha~Trinh, Newman Cheng, Joshua Bradley, Alex Chao, Apurva Mody, Steven Truitt, and Jonathan Larson.
\newblock From local to global: {A} graph {RAG} approach to query-focused summarization.
\newblock {\em CoRR}, abs/2404.16130, 2024.

\bibitem[\protect\citeauthoryear{He \bgroup \em et al.\egroup }{2024}]{G_Retriever}
Xiaoxin He, Yijun Tian, Yifei Sun, Nitesh~V. Chawla, Thomas Laurent, Yann LeCun, Xavier Bresson, and Bryan Hooi.
\newblock G-retriever: Retrieval-augmented generation for textual graph understanding and question answering.
\newblock {\em CoRR}, abs/2402.07630, 2024.

\bibitem[\protect\citeauthoryear{Ho \bgroup \em et al.\egroup }{2020}]{2WikiMultiHopQA}
Xanh Ho, Anh-Khoa Duong~Nguyen, Saku Sugawara, and Akiko Aizawa.
\newblock Constructing a multi-hop {QA} dataset for comprehensive evaluation of reasoning steps.
\newblock In Donia Scott, Nuria Bel, and Chengqing Zong, editors, {\em Proceedings of the 28th ICCL}, pages 6609--6625. ICCL, December 2020.

\bibitem[\protect\citeauthoryear{Jiang \bgroup \em et al.\egroup }{2023}]{jiang2023structgpt}
Jinhao Jiang, Kun Zhou, Zican Dong, Keming Ye, Xin Zhao, and Ji-Rong Wen.
\newblock Structgpt: A general framework for large language model to reason over structured data.
\newblock In {\em Proceedings of Conference on EMNLP 2023}, pages 9237--9251. ACL, 2023.

\bibitem[\protect\citeauthoryear{Kim \bgroup \em et al.\egroup }{2023}]{kim2023kg}
Jiho Kim, Yeonsu Kwon, Yohan Jo, and Edward Choi.
\newblock {KG-GPT:} {A} general framework for reasoning on knowledge graphs using large language models.
\newblock In {\em Findings of the EMNLP 2023}, pages 9410--9421. ACL, 2023.

\bibitem[\protect\citeauthoryear{Li \bgroup \em et al.\egroup }{2023}]{IE_GPT}
Bo~Li, Gexiang Fang, Yang Yang, Quansen Wang, Wei Ye, Wen Zhao, and Shikun Zhang.
\newblock Evaluating chatgpt's information extraction capabilities: An assessment of performance, explainability, calibration, and faithfulness.
\newblock {\em CoRR}, abs/2304.11633, 2023.

\bibitem[\protect\citeauthoryear{Liang \bgroup \em et al.\egroup }{2024}]{KAG}
Lei Liang, Mengshu Sun, Zhengke Gui, Zhongshu Zhu, Zhouyu Jiang, Ling Zhong, Yuan Qu, Peilong Zhao, Zhongpu Bo, Jin Yang, Huaidong Xiong, Lin Yuan, Jun Xu, Zaoyang Wang, Zhiqiang Zhang, Wen Zhang, Huajun Chen, Wenguang Chen, and Jun Zhou.
\newblock Kag: Boosting llms in professional domains via knowledge augmented generation.
\newblock {\em CoRR}, abs/2409.13731, 2024.

\bibitem[\protect\citeauthoryear{Munikoti \bgroup \em et al.\egroup }{2023}]{munikoti2023atlantic}
Sai Munikoti, Anurag Acharya, Sridevi Wagle, and Sameera Horawalavithana.
\newblock {ATLANTIC:} structure-aware retrieval-augmented language model for interdisciplinary science.
\newblock {\em CoRR}, abs/2311.12289, 2023.

\bibitem[\protect\citeauthoryear{Pan \bgroup \em et al.\egroup }{2024}]{roadmap}
Shirui Pan, Linhao Luo, Yufei Wang, Chen Chen, Jiapu Wang, and Xindong Wu.
\newblock Unifying large language models and knowledge graphs: {A} roadmap.
\newblock {\em {IEEE} Trans. Knowl. Data Eng.}, 36(7):3580--3599, 2024.

\bibitem[\protect\citeauthoryear{Panda \bgroup \em et al.\egroup }{2024}]{HOLMES}
Pranoy Panda, Ankush Agarwal, Chaitanya Devaguptapu, Manohar Kaul, and Prathosh~A P.
\newblock {HOLMES:} hyper-relational knowledge graphs for multi-hop question answering using llms.
\newblock In {\em Proceedings of 62nd Conference on ACL}, pages 13263--13282. ACL, 2024.

\bibitem[\protect\citeauthoryear{Paulheim}{2017}]{kg_survey}
Heiko Paulheim.
\newblock Knowledge graph refinement: {A} survey of approaches and evaluation methods.
\newblock {\em Semantic Web}, 8(3):489--508, 2017.

\bibitem[\protect\citeauthoryear{Peng \bgroup \em et al.\egroup }{2024}]{graph_review}
Boci Peng, Yun Zhu, Yongchao Liu, Xiaohe Bo, Haizhou Shi, Chuntao Hong, Yan Zhang, and Siliang Tang.
\newblock Graph retrieval-augmented generation: {A} survey.
\newblock {\em CoRR}, abs/2408.08921, 2024.

\bibitem[\protect\citeauthoryear{Soman \bgroup \em et al.\egroup }{2024}]{biomedical_kg_llm}
Karthik Soman, Peter~W Rose, John~H Morris, Rabia~E Akbas, Brett Smith, Braian Peetoom, Catalina Villouta-Reyes, Gabriel Cerono, Yongmei Shi, Angela Rizk-Jackson, et~al.
\newblock Biomedical knowledge graph-optimized prompt generation for large language models.
\newblock {\em Bioinformatics}, 40(9):btae560, 2024.

\bibitem[\protect\citeauthoryear{Sun \bgroup \em et al.\egroup }{2023}]{TOG}
Jiashuo Sun, Chengjin Xu, Lumingyuan Tang, Saizhuo Wang, Chen Lin, Yeyun Gong, Heung{-}Yeung Shum, and Jian Guo.
\newblock Think-on-graph: Deep and responsible reasoning of large language model with knowledge graph.
\newblock {\em CoRR}, abs/2307.07697, 2023.

\bibitem[\protect\citeauthoryear{Sun \bgroup \em et al.\egroup }{2024}]{sun2024docs2kg}
Qiang Sun, Yuanyi Luo, Wenxiao Zhang, Sirui Li, Jichunyang Li, Kai Niu, Xiangrui Kong, and Wei Liu.
\newblock Docs2kg: Unified knowledge graph construction from heterogeneous documents assisted by large language models.
\newblock {\em CoRR}, abs/2406.02962, 2024.

\bibitem[\protect\citeauthoryear{Taunk \bgroup \em et al.\egroup }{2023}]{taunk2023grapeqa}
Dhaval Taunk, Lakshya Khanna, Siri Venkata Pavan~Kumar Kandru, Vasudeva Varma, Charu Sharma, and Makarand Tapaswi.
\newblock Grapeqa: Graph augmentation and pruning to enhance question-answering.
\newblock In Ying Ding, Jie Tang, Juan~F. Sequeda, Lora Aroyo, Carlos Castillo, and Geert{-}Jan Houben, editors, {\em Companion Proceedings of the Conference on {WWW} 2023}, pages 1138--1144. {ACM}, 2023.

\bibitem[\protect\citeauthoryear{Trisedya \bgroup \em et al.\egroup }{2019}]{context_lost_edvi}
Bayu~Distiawan Trisedya, Jianzhong Qi, and Rui Zhang.
\newblock Entity alignment between knowledge graphs using attribute embeddings.
\newblock In {\em Proceedings of Conference on AAAI}, volume~33, pages 297--304, 2019.

\bibitem[\protect\citeauthoryear{Trivedi \bgroup \em et al.\egroup }{2022}]{musique}
Harsh Trivedi, Niranjan Balasubramanian, Tushar Khot, and Ashish Sabharwal.
\newblock Musique: Multihop questions via single-hop question composition.
\newblock {\em Trans. Assoc. Comput. Linguistics}, 10:539--554, 2022.

\bibitem[\protect\citeauthoryear{Wei \bgroup \em et al.\egroup }{2022}]{wei2022chain}
Jason Wei, Xuezhi Wang, Dale Schuurmans, Maarten Bosma, Brian Ichter, Fei Xia, Ed~H. Chi, Quoc~V. Le, and Denny Zhou.
\newblock Chain-of-thought prompting elicits reasoning in large language models.
\newblock In {\em Proceedings of the 2022 Conference of NeurIPS}, 2022.

\bibitem[\protect\citeauthoryear{Xu \bgroup \em et al.\egroup }{2024a}]{context_loss_kg}
Chengjin Xu, Muzhi Li, Cehao Yang, Xuhui Jiang, Lumingyuan Tang, Yiyan Qi, and Jian Guo.
\newblock Move beyond triples: Contextual knowledge graph representation and reasoning.
\newblock {\em arXiv preprint arXiv:2406.11160}, 2024.

\bibitem[\protect\citeauthoryear{Xu \bgroup \em et al.\egroup }{2024b}]{IE_llm}
Derong Xu, Wei Chen, Wenjun Peng, Chao Zhang, Tong Xu, Xiangyu Zhao, Xian Wu, Yefeng Zheng, Yang Wang, and Enhong Chen.
\newblock Large language models for generative information extraction: a survey.
\newblock {\em Frontiers Comput. Sci.}, 18(6):186357, 2024.

\bibitem[\protect\citeauthoryear{Yang \bgroup \em et al.\egroup }{2018}]{HotpotQA}
Zhilin Yang, Peng Qi, Saizheng Zhang, Yoshua Bengio, William~W. Cohen, Ruslan Salakhutdinov, and Christopher~D. Manning.
\newblock Hotpotqa: {A} dataset for diverse, explainable multi-hop question answering.
\newblock In {\em Proceedings of the Conference on EMNLP 2018}, pages 2369--2380. ACL, 2018.

\bibitem[\protect\citeauthoryear{Yang \bgroup \em et al.\egroup }{2024}]{fact_checking_with_kg}
Linyao Yang, Hongyang Chen, Zhao Li, Xiao Ding, and Xindong Wu.
\newblock Give us the facts: Enhancing large language models with knowledge graphs for fact-aware language modeling.
\newblock {\em {IEEE} Trans. Knowl. Data Eng.}, 36(7):3091--3110, 2024.

\bibitem[\protect\citeauthoryear{Yasunaga \bgroup \em et al.\egroup }{2021}]{yasunaga2021qa}
Michihiro Yasunaga, Hongyu Ren, Antoine Bosselut, Percy Liang, and Jure Leskovec.
\newblock {QA-GNN:} reasoning with language models and knowledge graphs for question answering.
\newblock In {\em Proceedings of the 2021 Conference of NAACL}, pages 535--546. ACL, 2021.

\bibitem[\protect\citeauthoryear{Zhang and Soh}{2024}]{kg_extract_llm}
Bowen Zhang and Harold Soh.
\newblock Extract, define, canonicalize: An llm-based framework for knowledge graph construction.
\newblock In {\em Proceedings of the 2024 Conference on EMNLP}, pages 9820--9836. ACL, 2024.

\bibitem[\protect\citeauthoryear{Zhang \bgroup \em et al.\egroup }{2024}]{KnowGPT}
Qinggang Zhang, Junnan Dong, Hao Chen, Daochen Zha, Zailiang Yu, and Xiao Huang.
\newblock Knowgpt: Knowledge graph based prompting for large language models.
\newblock In Amir Globersons, Lester Mackey, Danielle Belgrave, Angela Fan, Ulrich Paquet, Jakub~M. Tomczak, and Cheng Zhang, editors, {\em In Proceedings of Conference on NeurIPS 2024}, 2024.

\bibitem[\protect\citeauthoryear{Zhang \bgroup \em et al.\egroup }{2025a}]{RATT}
Jinghan Zhang, Xiting Wang, Weijieying Ren, Lu~Jiang, Dongjie Wang, and Kunpeng Liu.
\newblock {RATT:} {A} thought structure for coherent and correct {LLM} reasoning.
\newblock In {\em Proceedings of 39th Conference on AAAI 2025}, pages 26733--26741. {AAAI} Press, 2025.

\bibitem[\protect\citeauthoryear{Zhang \bgroup \em et al.\egroup }{2025b}]{G_rag}
Qinggang Zhang, Shengyuan Chen, Yuanchen Bei, Zheng Yuan, Huachi Zhou, Zijin Hong, Junnan Dong, Hao Chen, Yi~Chang, and Xiao Huang.
\newblock A survey of graph retrieval-augmented generation for customized large language models.
\newblock {\em CoRR}, abs/2501.13958, 2025.

\bibitem[\protect\citeauthoryear{Zhong \bgroup \em et al.\egroup }{2024}]{kgc}
Lingfeng Zhong, Jia Wu, Qian Li, Hao Peng, and Xindong Wu.
\newblock A comprehensive survey on automatic knowledge graph construction.
\newblock {\em {ACM} Comput. Surv.}, 56(4):94:1--94:62, 2024.

\bibitem[\protect\citeauthoryear{Zhu \bgroup \em et al.\egroup }{2024}]{llm_for_kg}
Yuqi Zhu, Xiaohan Wang, Jing Chen, Shuofei Qiao, Yixin Ou, Yunzhi Yao, Shumin Deng, Huajun Chen, and Ningyu Zhang.
\newblock Llms for knowledge graph construction and reasoning: recent capabilities and future opportunities.
\newblock {\em World Wide Web {(WWW)}}, 27(5):58, 2024.

\end{thebibliography}

\onecolumn
\section{Appendix}

\tcbset{
  colback=gray!12,
  colframe=black,
  width=0.9\textwidth,
  coltitle=white,
  fonttitle=\bfseries\large,
  colupper=black,
  boxrule=0.8pt,
  arc=3mm,
  top=2mm, bottom=2mm,
  left=3mm, right=3mm,
  breakable,
  title={}
}

\begin{center}
   \begin{tcolorbox}[title=Triples Extraction Prompt]
\begin{lstlisting}[basicstyle=\small, breaklines=true, breakindent=0pt, linewidth=\columnwidth, columns=fullflexible ]
--Objective--
Analyze the given text to extract structured information about entities and their relationships with strict ID consistency.
--Entity Extraction Rules--
1. Entity Identification:
   - Assign sequential EIDs starting from E1 (E1, E2, E3,...)
   - Maintain exact case from source text
   - Merge references to the same entity before assigning EIDs, and prohibit the use of non-explicit pronouns such as he, she, the film, etc.
   - Format entities as:
     [entity | EID | Type | "Entity Name" | Description]
2. Entity Requirements:
    - Types can include a wide range of categories such as People, Organizations, Locations, Events, Time Periods, Products, Concepts, General Entities, Event Entities, etc.
    - Type must use specific natural categories (e.g., "Medical Device" not "PRODUCT")
    - Include functional context in description
--Relationship Extraction Rules--
1. Validation Requirements:
   - Verify existence of both EIDs in entity list
   - Prohibit relations with unregistered EIDs
   - Require direct textual evidence
2. Format Enforcement:
   - Relation format: [relation | SourceEID | RelationType | TargetEID | "exact quote"]
   - Block relations where EID gap > current entity count
--Consistency Checks--
1. ID Validation:
   - Entity EIDs must form unbroken sequence
   - Prohibit duplicate EIDs
   - Restrict EID creation to entity section
2. Cross-reference Validation:
   - Relationship EIDs must match existing entity EIDs
   - Require bi-directional EID verification
--Error Prevention Measures--
1. ID Generation Protocol:
   Relation IDs prohibited
   String-to-EID conversion required
2. Processing Order:
   1. Full entity extraction
   2. Coreference resolution
   3. Relationship validation
   4. Final output assembly
--Strict Enforcement--
- Reject relations with invalid EIDs
- Terminate processing on EID mismatch
- Require entity-relation EID parity
- Prohibit special characters in EIDs
--Text Input--
"{input_text}"
Output:
\end{lstlisting}
\end{tcolorbox}

\clearpage

\begin{tcolorbox}[title=Reasoning Prompt]
\centering
\begin{lstlisting}[basicstyle=\small, breaklines=true, breakindent=0pt, linewidth=\columnwidth, columns=fullflexible ]
--Goal--  
Given a question and retrieved knowledge graph triplets (entity, relation, entity) along with text information, you must answer the question through logical, step-by-step reasoning. Prioritize using **only the provided contextual information** first. If insufficient data is available in the context, supplement with **internal knowledge** clearly marked. Structure your response into two sections as follows:  
### **Structure Requirements**  
1. Reasoning Process:  
   - Use **numbered steps** with **bold headings** (e.g., **Step 1: Verify Contextual Data**).  
   - Explicitly state whether information is derived from **context** or **internal knowledge**.  
2. Final Answer:  
   - Provide a concise, direct phrase without explanation.
--Attentions--
 - The output must strictly follow the format in the Examples.
 - The final answer must be concise.
--Examples-- 
**Example 1**  
Contextual Information:
Triples:
<Inception | released in | 2010>  
<Inception | genre | science fiction>  
<Inception | directed by | Christopher Nolan>  
<Interstellar | directed by | Christopher Nolan>  
<Interstellar | released in | 2014>
Related Text:
"Inception" was released in 2010.
Christopher Nolan directed both "Inception" and "Interstellar."
Both "Inception" and "Interstellar" are famous films directed by Christopher Nolan.
Question:  
"Which director of a 2010 sci-fi movie also directed a film released in 2014?"  
Output:  
Reasoning Process:  
1. **Identify 2010 sci-fi movies in context**:  
   - From context -> Inception qualifies (released in 2010, genre science fiction).  
2. **Find the director of Inception from context**:  
   - Context states director is Christopher Nolan.  
3. **Check context for 2014 films directed by Christopher Nolan**:  
   - Check 2014 films in context: Interstellar (2014, directed by Christopher Nolan).
Final Answer:  
Christopher Nolan   
--Real Data--
Contextual Information:  
{text}  

Question:
{question}  

Output:  
\end{lstlisting}
\end{tcolorbox}

\centering
\clearpage
\begin{tcolorbox}[title=Missing Knowledge Identification Prompt]
\begin{lstlisting}[basicstyle=\small, breaklines=true, breakindent=0pt, linewidth=\columnwidth, columns=fullflexible ]
--Goal--
Given the provided contextual information and the question, analyze what specific information is missing-but necessary or helpful-to fully, accurately, and contextually answer the question.
Then, express each missing piece of information as an independent, atomic question.
--Instructions--
- Carefully review the Contextual Information (including Triples and Related Text) and ensure that you only generate questions for information not already explicitly provided, but still required or useful to answer the original question comprehensively.
- In addition to the core facts directly required by the question, also consider any relevant details, background, context, definitions, conditions, processes, or clarifications that would make the answer more complete or informative.
- Each generated question should address one clear, specific unknown, phrased as independently and atomically as possible.
- If multiple facts or aspects are needed for a thorough answer, break them down into the smallest possible units-each question should target only one fact, attribute, or relationship.
- Avoid redundancy; do not ask about information already present in the Contextual Information.
- Output only the missing questions; do not include any explanations or extra content.
--Example 1--
Contextual Information:
Triples:
<Albert Einstein | born in | Ulm>
<Albert Einstein | born on | 14 March 1879>
<Albert Einstein | known for | theory of relativity>
<Albert Einstein | won Nobel Prize in | 1921>
<1921 Nobel Prize in Physics | awarded to | Albert Einstein>
Question: "For what did Albert Einstein receive the Nobel Prize in Physics?"
Output:
What was the 1921 Nobel Prize in Physics awarded for?
What was the official reason or citation for awarding Albert Einstein the 1921 Nobel Prize in Physics?
Who were the other nominees for the 1921 Nobel Prize in Physics?
What were the criteria for awarding the Nobel Prize in Physics in 1921?
--Example 2--
Contextual Information:
Triples:
<The Moon | orbits | Earth>
<Earth | part of | Solar System>
<Solar System | includes | Sun>
Question: "How long does it take for the Moon to complete one orbit around the Earth, and how far is the Moon from the Earth?"
Output:
What is the orbital period of the Moon around the Earth?
What is the average distance from the Moon to the Earth?
What factors affect the Moon's orbital period?
What is the shape of the Moon's orbit around the Earth?
--Example 3--
Contextual Information:
Triples:
<Apple Inc. | founded by | Steve Jobs>
<Apple Inc. | founded on | April 1, 1976>
<Apple Inc. | headquarters in | Cupertino, California>
Question: "Where was Apple Inc. originally founded and who were its founders?"
Output:
Where was Apple Inc. originally founded?
Who, besides Steve Jobs, founded Apple Inc.? 
What was the original name of Apple Inc. at the time of founding?
What were the circumstances or motivations behind the founding of Apple Inc.?
--Real Data--
Contextual Information:
{context_info}
Question: "{question}"
Output: 
\end{lstlisting}
\end{tcolorbox}

\clearpage
\begin{tcolorbox}[title=Knowledge Graph Enrichment Prompt]
\begin{lstlisting}[basicstyle=\small, breaklines=true, breakindent=0pt, linewidth=\columnwidth, columns=fullflexible ]
--Goal--
Given the provided Contextual Information-including Existing Triples, Sub-questions, and Related Sentences-extract all new and relevant triples from the Related Sentences that (1) directly provide information necessary to answer the Sub-questions or add meaningful context, and (2) are not already present in the Existing Triples. The output must strictly follow the format and requirements below.
--Detailed Instructions--
Analyze the Sub-questions: Carefully read all Sub-questions to identify the specific information required, as well as any supporting or contextual details that would help answer them.
Locate Supporting Information: Extract information found directly in the "Related Sentences" that is relevant to the Sub-questions, including both direct answers and closely related facts or context.
Triple Extraction Criteria:
Extract any information that directly supports or adds relevant context to the Sub-questions.
Extracted relations must not already exist in the Existing Triples.
Each triple must have clear and explicit evidence from the Related Sentences (i.e., a supporting phrase from the original text).
Do not infer, expand, or merge information-extract only what is explicitly stated or unambiguously implied.
Ensure entity definitions are consistent and unambiguous (e.g., person names, locations, organizations).
Entities may be extracted multiple times if necessary (i.e., repeated entity extraction is allowed for clarity or completeness).
Relations must not be duplicated-each relation should be unique in the output.
If a sentence provides multiple relevant details, extract each as a separate triple.
Output Requirements:
Output only Entities and Relations, strictly following the format below.
Do not include explanations, comments, or any extra content.
Before outputting, ensure all new relations are not duplicated in the Existing Triples or within your output.
Every EntityID (EID) used in Relations must be present and defined in Entities.
In Relations, both SubjectEntityID and ObjectEntityID must be EIDs, not entity names.
--Output Format--
Entities:
[entity | EntityID | Type | "Name" | Description]
Relations:
[relation | SubjectEntityID | Predicate | ObjectEntityID | "Evidence or supporting phrase"]
--Example 1--
Existing Triples:
<Marie Curie | won Nobel Prize in Physics | 1903>
<Marie Curie | won Nobel Prize in Chemistry | 1911>
Sub-questions:
When did Marie Curie win her Nobel Prizes?
Who did Marie Curie share the 1903 Nobel Prize in Physics with?
Related Sentences:
Nobel Prizes | Marie Curie shared the 1903 Nobel Prize in Physics with Pierre Curie and Henri Becquerel.
Output:
Entities:
[entity | E1 | Person | "Marie Curie" | Physicist and chemist]
[entity | E2 | Person | "Pierre Curie" | Physicist]
[entity | E3 | Person | "Henri Becquerel" | Physicist]
Relations:
[relation | E1 | shared Nobel Prize with | E2 | "shared the 1903 Nobel Prize in Physics with Pierre Curie"]
[relation | E1 | shared Nobel Prize with | E3 | "shared the 1903 Nobel Prize in Physics with Henri Becquerel"]
--Attention--
For every EntityID (EID) used in Relations, ensure the same EID is present and defined in Entities.
In Relations, both SubjectEntityID and ObjectEntityID must be E<IDs> format (not entity names).
--Real Data--
Existing Triples:
{context_info}
Sub-questions:
{sub_questions}
Related Sentences:
{context}
Output:
\end{lstlisting}
\end{tcolorbox}
 \end{center}

\end{document}